# Predicting optical coherence tomography-derived diabetic macular edema grades from fundus photographs using deep learning

Avinash V Varadarajan[1*]
Pinal Bavishi[1*]
Paisan Raumviboonsuk, MD[2*]
Peranut Chotcomwongse, MD[2†]
Subhashini Venugopalan, PhD[1†]
Arunachalam Narayanaswamy, PhD[1†]
Jorge Cuadros, OD, PhD[3†]
Kuniyoshi Kanai, OD, PhD[4]
George Bresnick, MD[3]
Mongkol Tadarati, MD[2]
Sukhum Silpa-archa, MD[2]
Jirawut Limwattanayingyong, MD[2]
Variya Nganthavee, MD[2]
Joe Ledsam, MBChB[5]
Pearse A Keane, MD[6]
Greg S Corrado, PhD[1]
Lily Peng, MD, PhD[1‡**]
Dale R Webster, PhD[1‡]

[1]Google AI, Google, Mountain View, CA, USA
[2]Department of Ophthalmology, Faculty of Medicine, Rajavithi Hospital, College of Medicine, Rangsit University, Bangkok, Thailand
[3]EyePACS LLC, Santa Cruz, CA, USA
[4]University of California, Meredith Morgan Eye Center, Berkeley, CA, USA
[5]Google DeepMind, Google, London, UK
[6]NIHR Biomedical Research Centre for Ophthalmology, Moorfields Eye Hospital NHS Foundation Trust and UCL Institute of Ophthalmology, United Kingdom

[*]Equal Contribution
[†]Equal Contribution
[‡]Equal Contribution
[**]Corresponding Author

**Abstract**

Diabetic eye disease is one of the fastest growing causes of preventable blindness. With the advent of anti-VEGF (vascular endothelial growth factor) therapies, it has become increasingly important to detect center-involved diabetic macular edema (ci-DME). However, center-involved diabetic macular edema is diagnosed using optical coherence tomography (OCT), which is not generally available at screening sites because of cost and workflow constraints. Instead, screening programs rely on the detection of hard exudates in color fundus photographs as a proxy for DME, often resulting in high false positive or false negative calls. To improve the accuracy of DME screening and enable more timely referrals for these patients, we trained a deep learning model to use color fundus photographs to predict ci-DME. Our model had an ROC-AUC of 0.89 (95% CI: 0.87-0.91), which corresponds to a sensitivity of 85% at a specificity of 80%. In comparison, three retinal specialists had similar sensitivities (82-85%), but only half the specificity (45-50%, p<0.001 for each comparison with model). The positive predictive value (PPV) of the model was 61% (95% CI: 56-66%), approximately double the 36-38% by the retinal specialists. In addition to predicting ci-DME, our model was able to detect the presence of intraretinal fluid with an AUC of 0.81 (95% CI: 0.81-0.86) and subretinal fluid with an AUC of 0.88 (95% CI: 0.85-0.91). The ability of deep learning algorithms to make clinically relevant predictions that generally require sophisticated 3D-imaging equipment from simple 2D images has broad relevance to many other applications in medical imaging.

**Introduction**

Diabetic macular edema (DME) is a late stage of diabetic eye disease that is characterized by retinal thickening in the macula, often accompanied by hard exudate deposition, and resultant vision loss. It is one of the most common reasons for referrals to diabetic eye clinics and affects 3-33% of patients with diabetes.[1] The wide range of prevalences reflects the varied bases for defining the condition as well as the varied composition of the populations studied. Currently, the first-line treatment for DME is anti-vascular endothelial growth factor (anti-VEGF) agents.[2–4] To determine eligibility for anti-VEGF treatment of DME, most of the major clinical trials measured macular thickening using optical coherence tomography (OCT), and initiated treatment if a patient met the criteria for a particular type of DME.[5,6] This type of DME is now commonly called center-involved DME (ci-DME) in clinical practice. As such, findings on OCT along with impaired visual acuity has become a widely-accepted standard of care for determining DME treatment.[7]

However, despite improvements in therapy, the detection of ci-DME remains a challenge because adding OCTs to the screening process is too costly and logistically difficult to implement widely. Globally, there are 425 million patients with diabetes[8] and most clinical guidelines recommend that all of them are screened annually.[9] Currently, selection of patients who may meet treatment criteria is performed during these screenings, which typically utilize monoscopic fundus images. These images are then evaluated for the presence of hard exudates within one optic disc diameter of the center of the macula, a proxy for ci-DME.[10] However, this proxy was developed based on an older standard of care, and some studies have shown that hard exudates have both poor positive predictive value and poor sensitivity for ci-DME. MacKenzie *et*

*al.* reported that only 42% of patients with hard exudates were found to have DME on OCT,[11] and Wang *et al.* reported that a third of patient-eyes with DME detected on OCTs lacked features such as hard exudates on monoscopic fundus photographs.[12] Wong *et al* reported a false positive rate of 86.6% for DME screening with existing strategies.[13] As such, the potential of DR screening and timely referral for DME is handicapped by an inability to reliably detect ci-DME via human-evaluation of fundus photographs alone.

A potential solution lies in the use of deep learning algorithms, which have been applied to a variety of medical image classification tasks,[14–16] including for retinal imaging.[17–20] Encouragingly, in addition to achieving expert-level performance for grading fundus images, deep learning algorithms are able to make predictions for which the underlying association with fundus images were previously unknown, such as cardiovascular risk factors[21] and refractive error.[22] Thus, we hypothesized that deep learning could be leveraged to directly predict the OCT-derived DME grade using monoscopic fundus photographs.

**Results**

To leverage deep learning as a potential solution to reliably detect ci-DME, we propose developing models on fundus photographs, but using ci-DME diagnoses derived from expert inspection of OCT (Figure 1). To train and validate the model, cases were gathered retrospectively from the Rajavithi Hospital in Bangkok, Thailand. Because these cases were gathered from those referred into the retina clinic for further evaluation, the disease distribution is consistent with a population presenting to specialty clinics, and enriched for more severe disease as compared to a DR screening population. Details of the development and clinical validation datasets are presented in Table 1. The development dataset consisted of 6,039 images from 4,035 patients and the primary clinical validation dataset consisted of 1033 images from 697 patients. For some patients, only one eye was included because the fellow eye fell under the exclusion criteria. ci-DME was conservatively defined as center point thickness >= 250 μm measured via manual caliper measurements excluding the retinal pigment epithelium.[23,24] We trained a model using this development dataset to predict ci-DME using fundus photographs as input.

Our model showed a higher performance in detecting cases with and without ci-DME from monoscopic fundus images compared to manual grading of fundus images (Table 2 and Figure 2). For ci-DME, the model had a sensitivity of 85% at a specificity of 80%. Three retinal specialists had sensitivities ranging from 82-85% at specificities ranging from 45-50%. The performance improvements held true even if other common criteria for calling DME for monoscopic images were used (Figure S1), such as changing the definition of DME based on the location of the hard exudates. Additional analyses were also performed at other thickness thresholds for ci-DME at center point thickness >=280 μm and >=300 μm, which showed similar or better results compared to the conservative >=250 μm cut off point without model retraining (Figure S2). When compared to manual grading, our model had a 30-35% absolute higher specificity at the same sensitivity ($p<0.001$ for comparison with each retinal specialist). When matched to have the same specificity, the model had a 11-14% absolute higher sensitivity (96% vs. 82-85%, $p<0.001$ for all comparisons).

In addition to predicting ci-DME, our model was able to predict presence of intraretinal and subretinal fluid. Our model had an AUC of 0.81 (95% CI: 0.81-0.86) for detecting intraretinal fluid presence and an AUC of 0.88 (95% CI: 0.85-0.91) for subretinal fluid presence (Figure 3).

In addition to the primary clinical validation dataset, the model was also applied to a secondary validation dataset, EyePACS-DME, to examine the model's generalizability. This dataset consists of 990 images with moderate, severe non-proliferative DR or proliferative DR, a subset of data previously gathered during another DME study.[25] The images were gathered using a Canon CR-DGi camera and OCTs were taken with a Optovue iVue machine from a U.S.-based population (see methods). There are some notable differences in this dataset in comparison to the primary validation dataset, particularly in terms of defining and measuring ci-DME based on central subfield thickness and incorporation of inclusion/exclusion criteria (Table S2). Based on this different definition and inclusion criteria, the number of ci-DME cases in the secondary validation set was 7.8% compared to 27.2% in the primary clinical validation set. Thus, the model performance on the datasets cannot be compared directly in terms of absolute values (especially for metrics like PPV which depend a lot on the priori distribution). However, relative comparisons between the model and graders (in this instance EyePACS certified graders) can be drawn (Figure 4 and Table 3). Similar to the results of the primary validation, our model had a positive predictive value (PPV) roughly twice that of manual grading using hard exudates as proxy (35% [95% CI: 27%-44%] vs 18% [95% CI: 13%-23%]) and similar NPV (96% [95% CI: 95%-98%] vs 95% [95% CI: 94%-97%]). This translated to a similar sensitivity (57% [95% CI: 47%-69%] vs 55% [43%-66%]) but higher specificity (91% [95% CI: 89%-93%] vs 79% [95% CI: 76%-82%]).

Subsampling experiments, where new models were trained using titrated fractions of the dataset, showed that model performance continued to increase with larger training sets (see Figure 3 - where AUC increases with sample size). These results suggest that the accuracy of this prediction will likely continue to increase with dataset sizes larger than that in this study.

Figure 5 presents an analysis of the areas in the fundus image relevant for the model. When the model was trained on cropped fundus images containing only 0.25 optic disc diameter around the fovea (blue line), it achieved an AUC of 0.75. When it had access to 1.0 optic disc diameter around the fovea, the model achieved an AUC greater than 0.85, comparable with its performance on the full fundus image. However, the model trained on the region around the optic disc needed to see a lot more context (2.5 optic disc diameter) around the optic disc center to achieve an AUC exceeding 0.8. Based on these results, we believe the model primarily utilizes the area around the fovea to make ci-DME predictions.

**Discussion**

While the potential of deep learning to make novel predictions has been reported in literature,[21,22] this study is among the first robust examples of a model far exceeding expert performance for a task with high clinical relevance and potentially important implications for screening programs worldwide. The resultant model performed significantly better than retinal specialists for detecting ci-DME from fundus images in two datasets from very different populations. DME is the major cause of visual loss from DR. Prior to the use of anti-VEGF injections, the Early Treatment of Diabetic Retinopathy Study (ETDRS) showed that treatment of a subtype of DME with focal laser photocoagulation decreased the chance of vision loss.[26] Today, with anti-VEGF injections, the treatment of ci-DME can *improve* vision by approximately 10-13 letters as measured using the ETDRS visual acuity chart.[5] Anti-VEGF injections are now largely considered the gold standard of care with evidence that shows that delaying treatment of DME could lead to suboptimal visual improvement.[27] However, the current grading guidelines in screening programs were developed before the advent of anti-VEGF therapy and are not specifically designed for detecting ci-DME. The development of models that can better detect ci-DME in DR screening programs using existing equipment (color fundus cameras) is both scientifically interesting and clinically impactful.

For DR screening in particular, our model may lead to fewer false negatives for DME. Decreasing missed referrals for patients with ci-DME presenting with no hard exudates is a clear advantage of such a system. Visual acuity alone is not enough to rule out ci-DME as baseline characteristics from some well-known cohorts suggest that a substantial percentage of eyes with ci-DME still have good vision.[28,29] In addition, decreasing false positives is also important in resource-constrained settings. While many screening programs recommend closer follow up for patients with mild or worse DR, the urgency of follow up varies widely, especially in low resource settings. Per international guidelines (International Council of Ophthalmology Guidelines, American Academy of Ophthalmology), for patients with mild DR and no macular edema, referral is not always required and patients can be rescreened in 1-2 years in low/intermediate resource settings and 6-12 months in high resource settings. However, patients with suspected ci-DME need to be referred within a month. For patients with moderate nonproliferative DR and no macular edema, follow up changes from 6-12 months in low/intermediate resource or 3-6 months in high resource settings to 1 month (all resource settings) when there is ci-DME.[30,31] In this study, roughly 88% of the moderate non-proliferative patients from the EyePACS-DME dataset and 77% of those from the Thailand dataset who would have been referred urgently (and unnecessarily) using a hard exudate based referral criterion did not have ci-DME. Higher urgency referral of patients with moderate non-proliferative DR without DME (but presenting hard exudates) can be a major issue where there are limited resources for evaluation and treatment.

Furthermore, the center point thickness distribution of the false positive and false negative instances is better for the model when compared to retina specialists (Figure S3). For the model, 28% of the false positives have thickness > 225 microns and 35% of the false negatives have thickness < 275 microns. In comparison, for retina specialists, only 20% of the false positives have thickness > 225 microns and 17% of the false negatives have thickness < 275 microns. This shows that a significantly larger fraction of the model false positives and negatives are borderline cases as compared to retina specialists. In addition, the new model seems to be able to detect the presence of intraretinal and/or subretinal fluid, both of which merit closer monitoring and possibly treatment.[32] The ability to detect these pathologies is also novel since this is not a task that doctors can do accurately from fundus images.

Although the performance of the models on the secondary dataset is lower than that of the models on the primary dataset, the performance of the human graders on the secondary dataset is proportionally lower as well. From the primary to secondary dataset, PPV of the model decreased from 61% to 35% whereas of the graders decreased from 37% to 18%; sensitivity of the model decreased from 85% to 57% whereas of the graders decreased from 84% to 55%.

However, NPV of the model increased from 93% to 96% whereas of the graders increased from 88% to 95%; specificity of the model increased from 80% to 91% whereas of the graders increased from 47% to 79%. These results reflect the inherent differences between the two datasets but still support the better performance of the model over graders on both datasets. While the models trained in this study are more accurate than manual grading, there is capacity for improvement. Given the results of the subsampling experiments, it is likely that the accuracy of the model may continue to increase with larger dataset sizes.

From a scientific point of view, this work demonstrates the potential of deep learning to enable diagnostics from inexpensive hardware, that was only previously possible from expensive equipment. It also lays the groundwork for understanding how the model makes these predictions. The explanation technique employed in this study indicated that the region around the fovea is more relevant than the region near the optic disc for DME prediction from fundus images. Future work could involve diving deeper into the features around this area that is picked up by deep learning but overlooked by retinal specialists.

In a small non-randomized study, Scott *et al.*[33] showed a beneficial effect of focal and grid laser for eyes without central involvement that meet an older criteria for treatment known as clinically significant macular edema, similar to the initial ETDRS findings. These patients need to be referred from a DR screening program for closer follow up. Our model does not evaluate such cases. To address this, one would include stereoscopic imaging in addition to OCT as ground truth to train model(s) to specifically identify these cases. While there is some evidence of generalization to a secondary dataset, the confidence intervals are wide and the criteria for ci-DME for the EyePACS-DME dataset were different from those of the Thailand dataset. Some of the performance metrics reported in this study such as positive predictive value (PPV) and negative predictive value (NPV) are relevant only to populations whose severity distribution is similar to that of this study (e.g. patients referred to specialist clinics). Further studies should validate the model on additional larger datasets from other settings, including screening settings from other regions or geographies. Future studies should also include better standardization for ci-DME and inclusion/exclusion criteria as well as sub-analysis of patients who were treated for DME. Moreover, additional data diversity such as the use of ci-DME labels derived from other OCT devices by other manufacturers should be included in future work. Since the model was trained using treatment-naive fundus images, training on multiple images per eye (including with stereo pairs), and on eyes that have been treated for DME in the past could lead to better model performance. While our cropping experiments (Figure 5) shows that the model looks at the region around the fovea for predicting ci-DME, future work could further explore interpretability of the model [33]. Lastly, future work could also include health economic analysis to study the cost-effectiveness of such an approach.

Nevertheless, this study demonstrates that deep learning can be leveraged to identify the presence of ci-DME using the cheaper and more-widely available fundus photograph, at an accuracy exceeding that of manual grading using expert-derived rules. Similar approaches could be particularly valuable for other medical images, such as using radiographs or low-dose computed tomography to detect conditions that would otherwise require more expensive imaging techniques that expose patients to higher radiation doses. Importantly, we also use crops around the fovea and optic disc to explain how the model is making these predictions, lending confidence that the predictions will generalize to new unseen datasets.

**Methods**

This study was approved by the Ethics Committees or Institutional Review Boards of hospitals or health centers where retinal images of patients with diabetes were used in this study, including the Rajavithi Hospital (Bangkok, Thailand), Alameda Health Service (Alameda, CA, USA), and the University of California, Berkeley (Berkeley, CA, USA). Patients gave informed consent allowing their retinal images to be used. This study was registered in the Thai Clinical Trials Registry, Registration Number TCTR20180818002.

*Datasets*

For algorithm development, 7,072 images were gathered retrospectively from diabetic patients presenting to the retina clinic at Rajavithi Hospital in Bangkok, Thailand from January 2010-February 2018. Only cases that were naive to treatment (both intravitreal injections and lasers) were included. Cases where macular lesions may have hyporeflective spaces on OCT, such as Macular Telangiectasia Type 2, may interfere with the diagnosis of DME, such as idiopathic epimacular membrane, macular edema from other causes, or proliferative DR with neovascular membrane affecting the macula, were excluded from analysis.

Retinal fundus images were obtained using Kowa color fundus camera (VX-10 model, Kowa, Aichi, Japan). A single macula-centered color fundus photograph per eye was used in the study. If available, imaging from both eyes were included. OCTs were obtained using the Heidelberg Spectralis OCT (Heidelberg Engineering GmbH, Germany) and thickness measurements were measured manually (see below for measurement procedures).

Of the 7,072 images in the dataset, 6,039 were used for development while 1,033 were set aside for clinical validation. All images from a patient was present in either in development or validation sets, but not both. Fundus photographs in the validation set were manually graded by U.S. board-certified retinal specialists to assess the presence and location of hard exudates (yes, no, ungradable, within 500 μm or 1 disc diameter or 2 disc diameter from the center of the macula) and focal laser scars. In addition, retinal specialists provided their best clinical judgement of the presence of DME that took into account all the pathology present in the image.

To study generalizability of the model, the algorithm was applied to another dataset, EyePACS-DME, which is a subset of data that had been previously gathered for another DME study.[25] This dataset consisted of 990 macula-centered images from 554 patients with at least moderate DR based on grading by certified EyePACS graders (to roughly match the population of those who would be presenting to a retina clinic). No other exclusion criteria were applied to this dataset (e.g. exclusion of epiretinal membrane, etc). Fundus images were taken with a Canon CR-DGi camera (Ōta, Tokyo, Japan) and OCTs were taken with a Optovue iVue machine (Fremont, CA, USA).

*Measurement and assessment of OCT scans*

For the Thailand dataset, central subfield thickness, the value representing the thickness of the center of the macula in clinical trials for DME,[24] was not available in all eyes in the developmental dataset; therefore center point thickness, which was found to have high correlation with the central subfield thickness,[23] was measured for each eye to represent the thickness of the center of the macula.

The center point thickness of an eye of a patient was manually measured on the axis of the OCT scan where there was a slight elevation of the ellipsoid zone and the gap between the photoreceptor layer outer segment tip and the ellipsoid zone was the widest, indicating the center of the fovea where the cone cell density is the highest. Manual measurement was conducted using the straight-line measurement vector available with the Spectralis Eye Explorer software. The vector was put perpendicular to the highly reflective band of retinal pigment epithelium with one side of the vector rested on the highly reflective line of cone outer segment tip and the other side on the internal limiting membrane. Retinal pigment epithelium thickness was not included in this measurement. Intraretinal fluid was defined as present when a cystoid space of hypo-reflectivity was found within 500 μm of the foveal center of any OCT scans of a patient. Subretinal fluid was defined as present when a space of hypo-reflectivity was found between the retina and retinal pigment epithelium within 500 μm of the foveal center of any OCT scans of a patient.

The measurement of center point thickness and the assessment of presence of intraretinal fluid and subretinal fluid were conducted by 2 medical doctors experienced in clinical research and supervised by retinal specialists. 5% of patients were randomly selected to confirm all three measurements by a retinal specialist with 20 years of post-certification experience.

Eyes were divided into cases of no ci-DME and ci-DME. ci-DME was conservatively defined as eyes with >= 250 μm center point thickness, excluding the retinal pigment epithelium based upon manual measurement.[23,24] In addition to ci-DME, we also trained the model in a multi-task fashion to predict subretinal fluid and intraretinal fluid (details below). While cases with subretinal fluid and intraretinal fluid were not strictly included in the criteria in the clinical anti-VEGF trials for DME, referral for follow-up is warranted for these cases.

For the EyePACS-DME dataset, the manufacturer's automated segmentation algorithm was used to measure central subfield thickness. A cut off of 300 μm central subfield thickness was used as the cut-off point for ci-DME based on machine-specific adjustments.[34] The presence of intraretinal and subretinal fluid were not available in this dataset.

*Model*

Our deep learning algorithm for predicting ci-DME was built using the methods described by Gulshan *et al.*,[17] using the Inception-v3[35] neural network architecture. Briefly, we used a convolutional neural network[36] to predict ci- DME (center point thickness >= 250 μm), subretinal fluid presence and intraretinal fluid presence in a multi-task fashion. The input to the neural network was a color fundus photograph, and the output was a real-valued number between 0 and 1 for each prediction, indicating its confidence. For other hyperparameter and training details see Figure S4.

The parameters of the neural network were determined by training it on the fundus images and OCT-derived ci-DME grades in the development dataset. Repeatedly, the model was given a fundus image with a known output as determined by a grader looking at the patient's corresponding OCT . The model predicted its confidence in the output, gradually adjusting its parameters over the course of the training process to become more accurate. Note that the model never sees the actual OCT image during training or validation.

*Evaluating the algorithm*

To evaluate the performance of the model, we used the receiver operating characteristic (ROC) curve and calculated the area under the curve (AUC). The performance of the retinal specialists was marked by points on this curve indicating their sensitivity and specificity (Figure 2). The same model was also evaluated at increasing thresholds of thickness for ci-DME, without retraining (Figure S2). By choosing an operating point on the ROC curve that makes the model's specificity match that of retinal specialists, we also evaluated the model using Sensitivity, Specificity, Positive Predictive Value, Negative Predictive Value, Accuracy and Cohen's Kappa score[37] (Table 2).

*Statistical Analysis*

To assess the statistical significance of these results, we used the non-parametric bootstrap procedure: from the validation set of $N$ patients, sample $N$ patients with replacement and evaluate the model on this sample. By repeating this sampling and evaluation 2,000 times, we obtain a distribution of the performance metric (e.g. AUC), and report the 2.5 and 97.5 percentiles as 95% confidence intervals. For statistical comparisons, the permutation test was used with 2,000 random permutations.[38]

*Model Explanation*

We performed two experiments to determine which regions in a fundus image are most informative of DME. We focussed on two regions, the macula and the optic disc. First, a group comprised of ophthalmologists and optometrists manually marked the fovea and disc for all images in the Thailand dataset. We then trained and evaluated our model looking only at the region that is within a factor of optic disc diameters around the fovea (or equivalently the optic disc) with the rest of the fundus blacked out. We trained and evaluated different models for different radii, increasing the area that the model looks at to understand the importance of these regions in making the prediction. (Figure 5)

**Acknowledgements**

Dr. Nitee Ratprasatporn, Dr. Withawat Sapthanakorn, Dr. Vorarit Jinaratana, Dr. Yun Liu, Dr. Naama Hammel, Ashish Bora, Anita Misra, Dr. Ali Zaidi, Dr. Courtney Crawford, Dr. Jesse Smith for their assistance in reviewing the manuscript, image grading, and data collection.

P.A.K. is supported by an NIHR Clinician Scientist Award (NIHR-CS-2014-14-023). The views expressed are those of the author and not necessarily those of the NHS, the NIHR or the United Kingdom (UK) Department of Health.

**Data availability**

Restrictions apply to the sharing of patient data that support the findings of this study. This data may be made available to qualified researchers upon ethical approvals from Rajavithi Hospital and EyePACS.

**Code availability**

Machine learning models were developed and deployed using standard model libraries and scripts in TensorFlow. Custom code was specific to our computing infrastructure and mainly used for data input/output and parallelization across computers.

**Competing interests**

The authors are employees of Google.

# Figures & Tables

| Characteristics | Thailand dataset | | EyePACS-DME dataset |
|---|---|---|---|
| | **Development Set** | **Primary Clinical Validation Set** | **Secondary Clinical Validation Set** |
| Number of Patients | 4035 | 697 | 554 |
| Number of Fundus Images | 6039 | 1033 | 990 |
| Camera used for Fundus Images | Kowa VX-10 | | Canon CR-DGi |
| OCT device used for determining ci-DME | Heidelberg Spectralis | | Optovue iVue |
| Age: Mean, years (SD) | 55.6 (10.8) $n$=6038 | 55.8 (10.8) $n$=1033 | 62.0 (9.8) $n$=990 |
| Gender (% male) | 60.8% $n$=6036 | 62.4% $n$=1031 | 50.1% $n$=990 |
| Central Retinal Thickness: Mean, µm (SD) | 263.8 (146.5) $n$=6039 | 258.4 (132.8) $n$=1033 | 254.4 (56.3) $n$=990 |
| ci-DME, Center Point Thickness >= 250um in Thailand dataset. Central Subfield Thickness >= 300um in the Eyepacs-DME dataset | 28.3% $n$=6039 | 27.2% $n$=1033 | 7.8% $n$=990 |
| Subretinal Fluid Presence | 15.7% $n$=6039 | 15.1% $n$=1033 | N/A |
| Intraretinal Fluid Presence | 45.5% $n$=6039 | 46.3% $n$=1033 | N/A |

**Table 1**: Baseline characteristics of the development and primary clinical validation datasets. Note that the difference between total n and subcategories is missing data (e.g. not all images had age or sex)

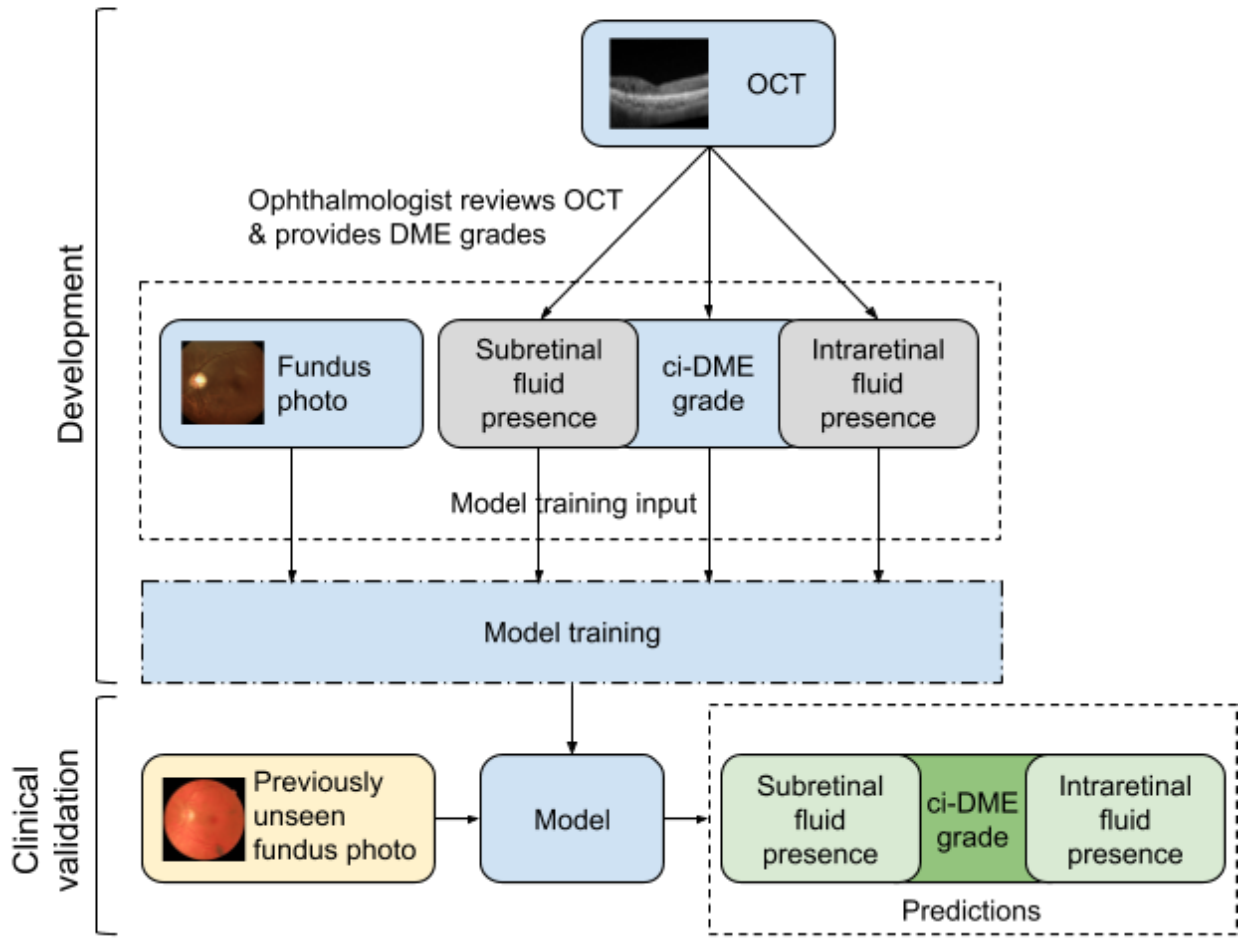

**Figure 1:** Illustration of our proposed approach for developing a ci-DME model. Ground truth for ci-DME were derived from a human grader analyzing the OCT for each case. Additionally subretinal fluid and intraretinal fluid presence grades were also collected. These ground truth labels and corresponding color fundus photos were used for model training. For clinical validation, the trained model takes in a new fundus photo and generates a predicted ci-DME grade, predicted subretinal fluid and intraretinal fluid presence grades.

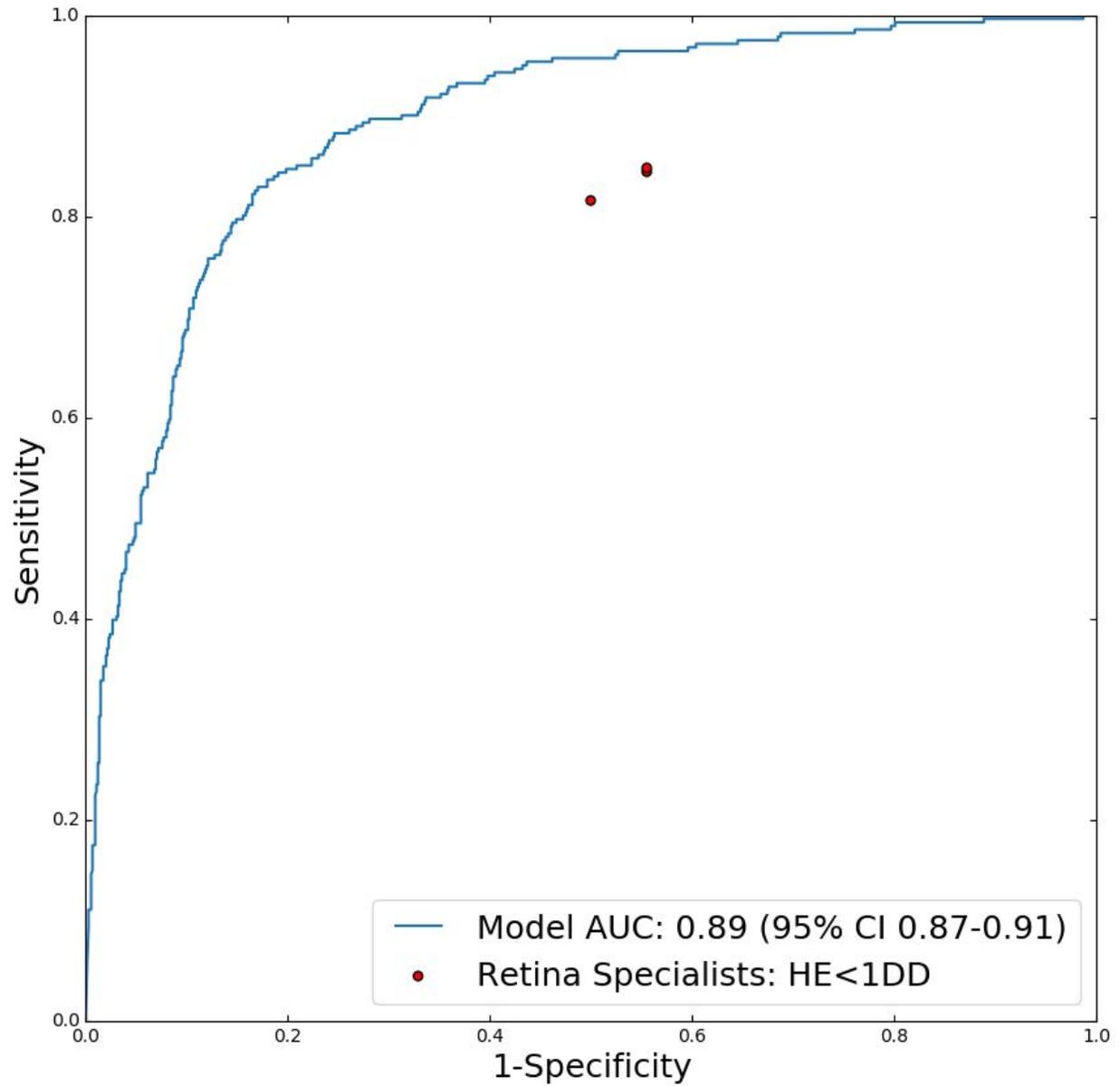

**Figure 2:** Receiver operating characteristic curve of the model with retinal specialists' grades shown as red dots for predicting ci-DME on the primary clinical validation set. All methods (i.e. the model and retinal specialists) rendered their grades using monoscopic fundus images only. The ground truth for ci-DME was derived using OCT (center point thickness>=250 μm).

| Metric | Model | Specialist 1 | Specialist 2 | Specialist 3 |
|---|---|---|---|---|
| Positive Predictive Value (%), 95% CI | 61% [56%-66%] $n$=1033 | 37% [33%-40%] $n$=1004 | 36% [33%-40%] $n$=987 | 38% [34%-42%] $n$=1001 |
| Negative Predictive Value (%), 95% CI | 93% [91%-95%] $n$=1033 | 88% [85%-91%] $n$=1004 | 89% [85%-92%] $n$=987 | 88% [84%-91%] $n$=1001 |
| Sensitivity (%), 95% CI | 85% [80%-89%] $n$=1033 | 84% [80%-89%] $n$=1004 | 85% [80%-89%] $n$=987 | 82% [77%-86%] $n$=1001 |
| Specificity (%), 95% CI | 80% [77%-82%] $n$=1033 | 45% [41%-48%] $n$=1004 | 45% [41%-48%] $n$=987 | 50% [47%-54%] $n$=1001 |
| Accuracy (%), 95% CI | 81% [79%-83%] n=1033 | 56% [52%-59%] n=1004 | 56% [52%-59%] n=987 | 59% [56%-62%] n=1001 |
| Cohen's Kappa, 95% CI | 0.57 [0.52-0.62] n=1033 | 0.21 [0.16-0.25] n=1004 | 0.21 [0.16-0.25] n=987 | 0.24 [0.19-0.28] n=1001 |

**Table 2:** Performance metrics (PPV, NPV, Sensitivity, Specificity, Accuracy and Cohen's Kappa) of the model for predicting ci-DME compared with 3 retinal specialists on the primary clinical validation set. For the model we chose an operating point that matched the sensitivity of the retinal specialists to calculate the metrics. The performance metrics for the model were calculated on the entire primary clinical validation set; for the retinal specialists it was calculated only on the images that they marked as gradable. Brackets denote 95% confidence intervals.

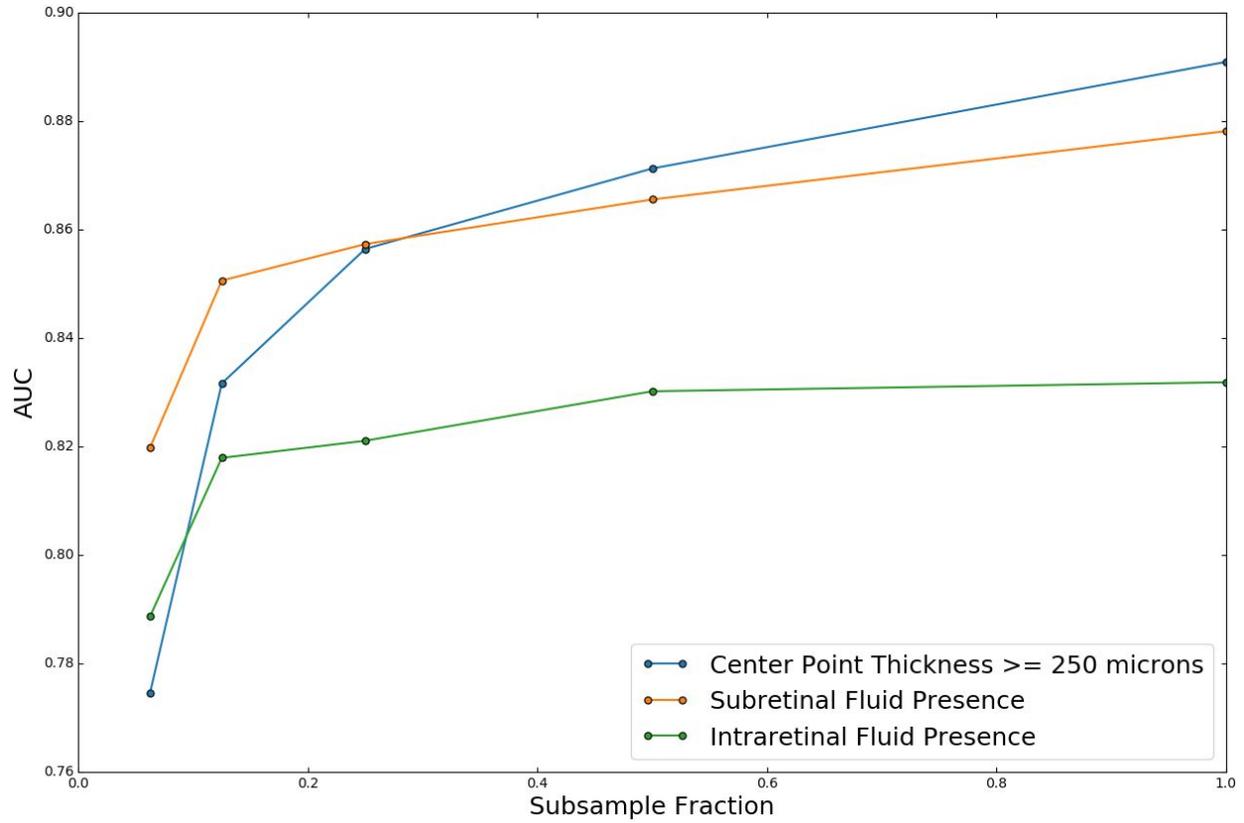

**Figure 3.** Effect of data size on predicting ci-DME, subretinal fluid and intraretinal fluid presence on the primary clinical validation set. A subsampled fraction of 1.0 indicates the entire training dataset. Model performance continues to increase with increased data suggesting that the accuracy of the predictions will likely improve if the model is trained with more data.

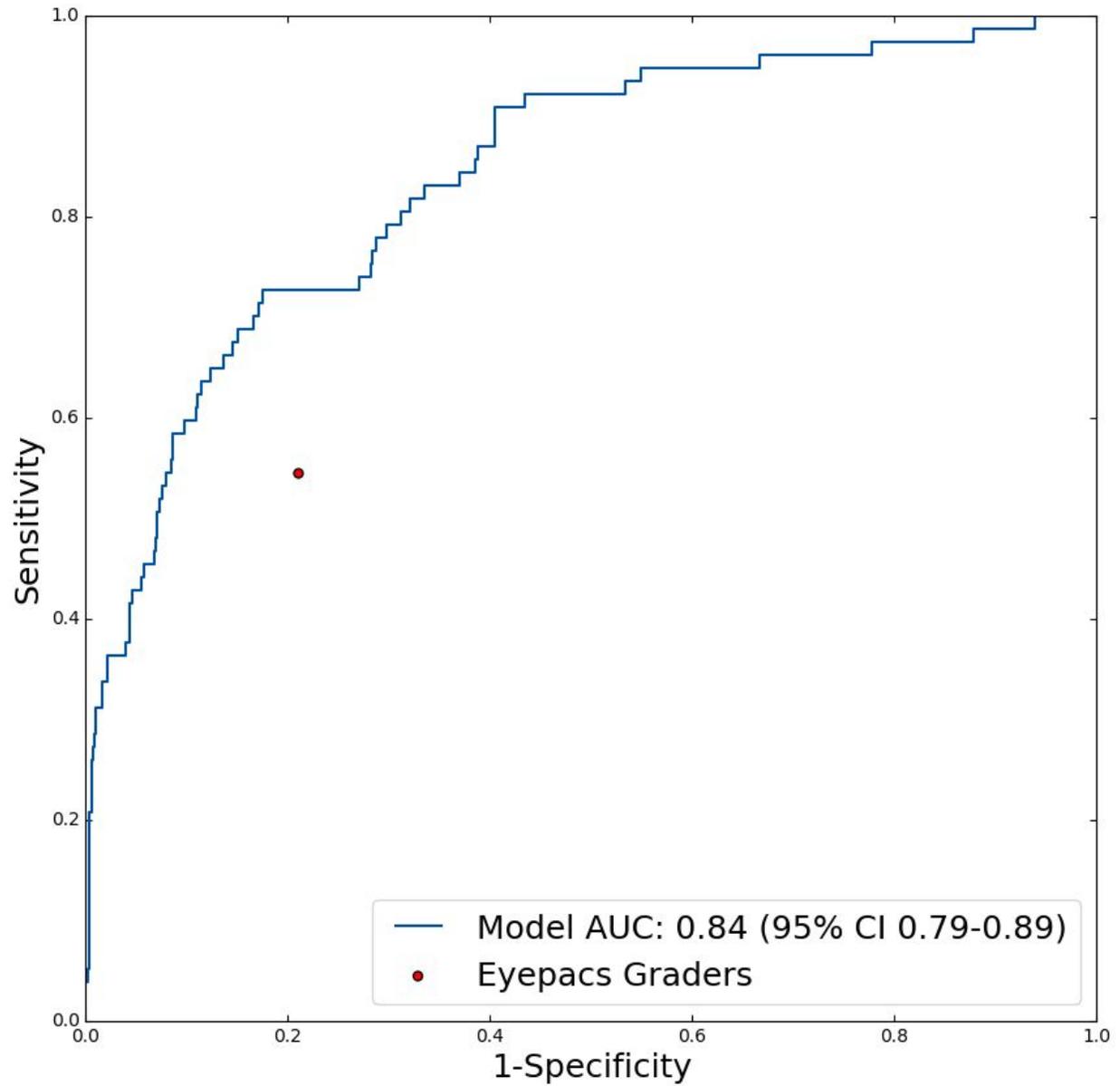

**Figure 4:** Receiver operating characteristic curve of the model with eyepacs graders' grades shown as a red dot for predicting ci-DME on the secondary clinical validation set. All methods (i.e. the model and eyepacs graders) rendered their grades using monoscopic fundus images only. The ground truth for ci-DME was derived using OCT (central subfield thickness>=300 μm).

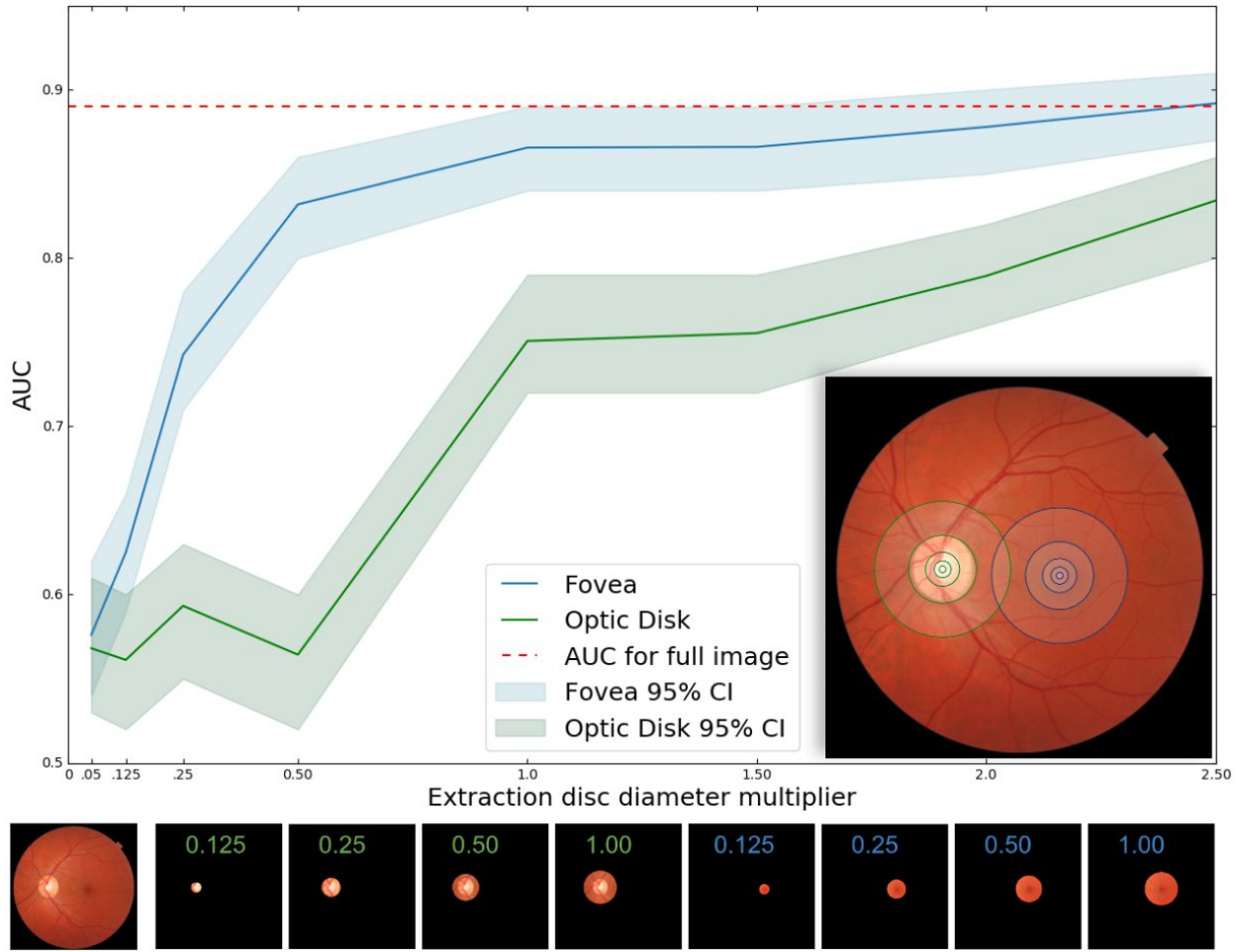

**Figure 5:** Plot showing model performance in predicting ci-DME on the primary clinical validation set, as measured by AUC when cropped circular images are used to train and validate the model. The blue line indicates the performance when cropped circular images of different sizes (radius of multiples of disc diameter from 0.05 to 2.5) centered at the fovea are used, while the green line indicates the corresponding performance when the crops are centered at the optic disc. (Inset) Image depicting some of the regions of different radii (0.05, 0.125, 0.25, 0.5, and 1 disc diameter) around the fovea and optic disc. (Bottom panel) Fundus image, followed by the crops extracted by centering at the optic disc (green) and fovea (blue), with the extraction radius in multiples of disc diameter indicated in each crop.

| Metric | Model | EyePACS Graders |
|---|---|---|
| Positive Predictive Value (%), 95% CI | 35% [27%-44%] | 18% [13%-23%] |
| Negative Predictive Value (%), 95% CI | 96% [95%-98%] | 95% [94%-97%] |
| Sensitivity (%), 95% CI | 57% [47%-69%] | 55% [43%-66%] |
| Specificity (%), 95% CI | 91% [89%-93%] | 79% [76%-82%] |
| Accuracy (%), 95% CI | 88% [86%-91%] | 77% [74%-80%] |
| Cohen's Kappa, 95% CI | 0.38 [0.29-0.47] | 0.17 [0.11-0.24] |

**Table 3:** Performance metrics (PPV, NPV, Sensitivity, Specificity, Accuracy and Cohen's Kappa) of the model for predicting ci-DME compared with eyepacs graders on the secondary clinical validation set

(n=990). For the model we chose an operating point that matched the sensitivity of the eyepacs graders to calculate the metrics. Brackets denote 95% confidence intervals.

**Supplement**

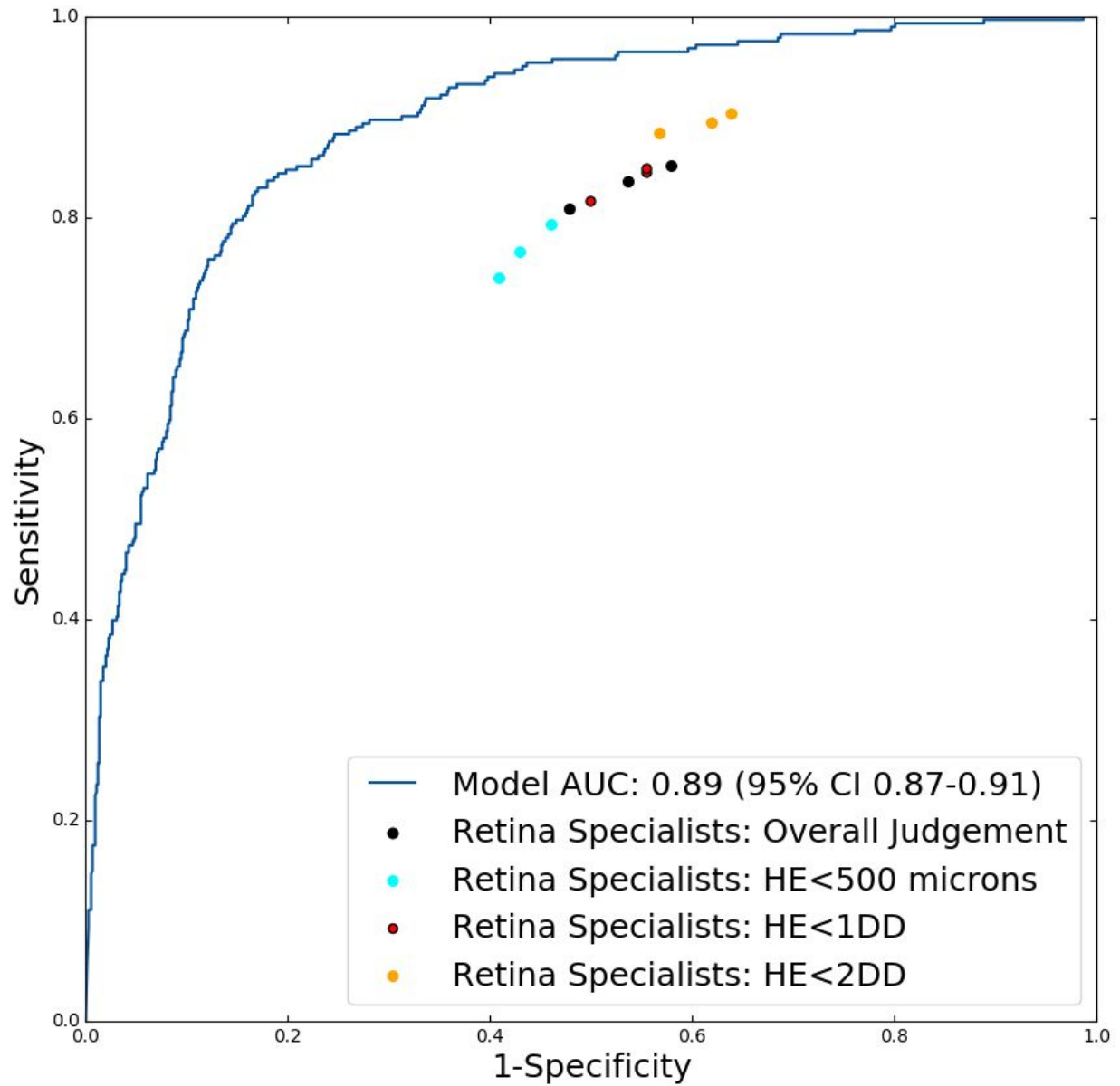

**Figure S1:** Receiver operating characteristic curve of the model with retinal specialists' grades shown as red, yellow, and cyan dots for predicting ci-DME using different criteria for manual grading for DME on the primary clinical validation set. All methods (i.e. the model and retinal specialists) rendered their grades using monoscopic fundus images only. The ground truth was derived using OCT (center point thickness>=250 μm).

A. CPT>=280 μm

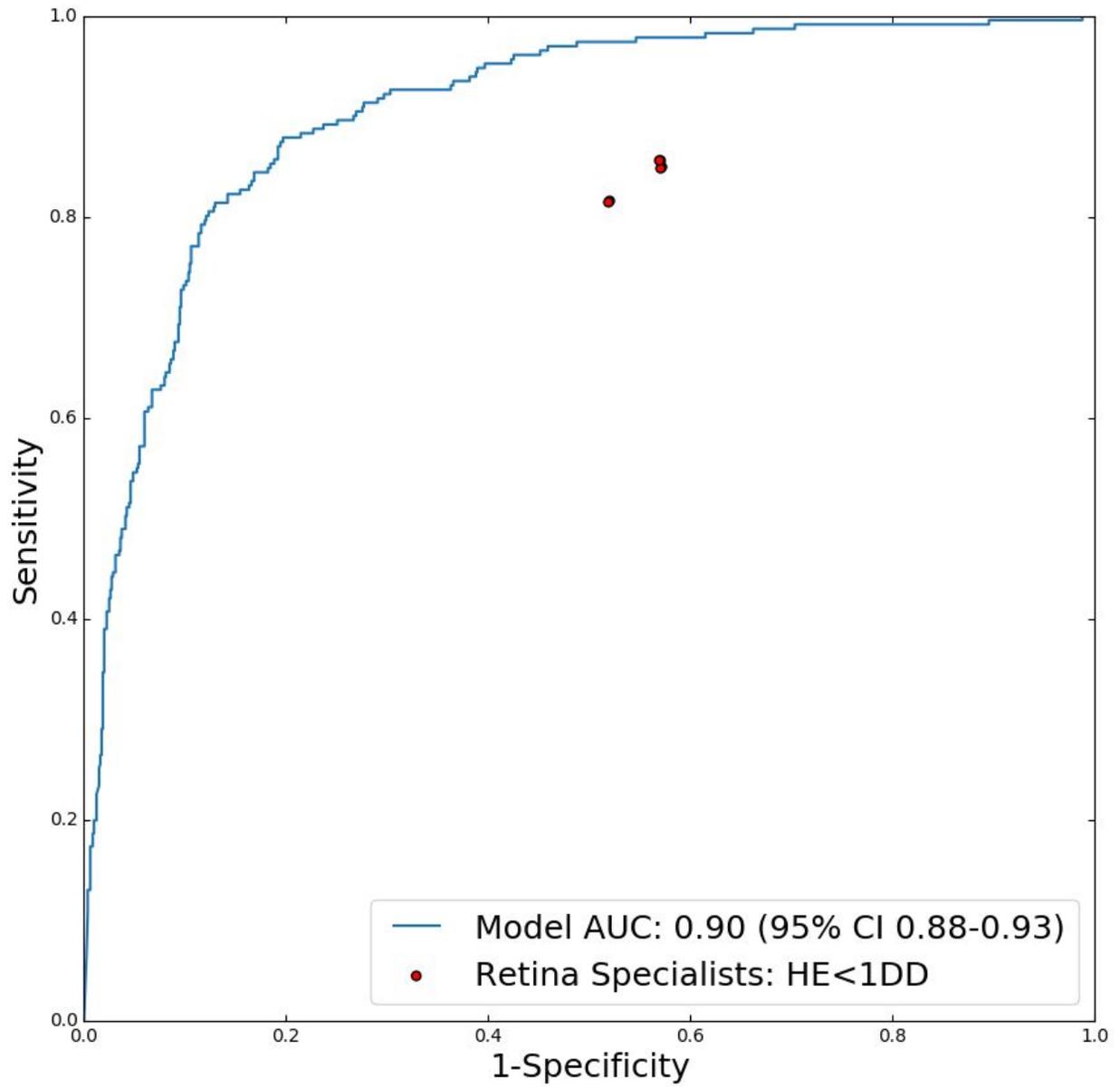

**B. CPT>=300 μm**

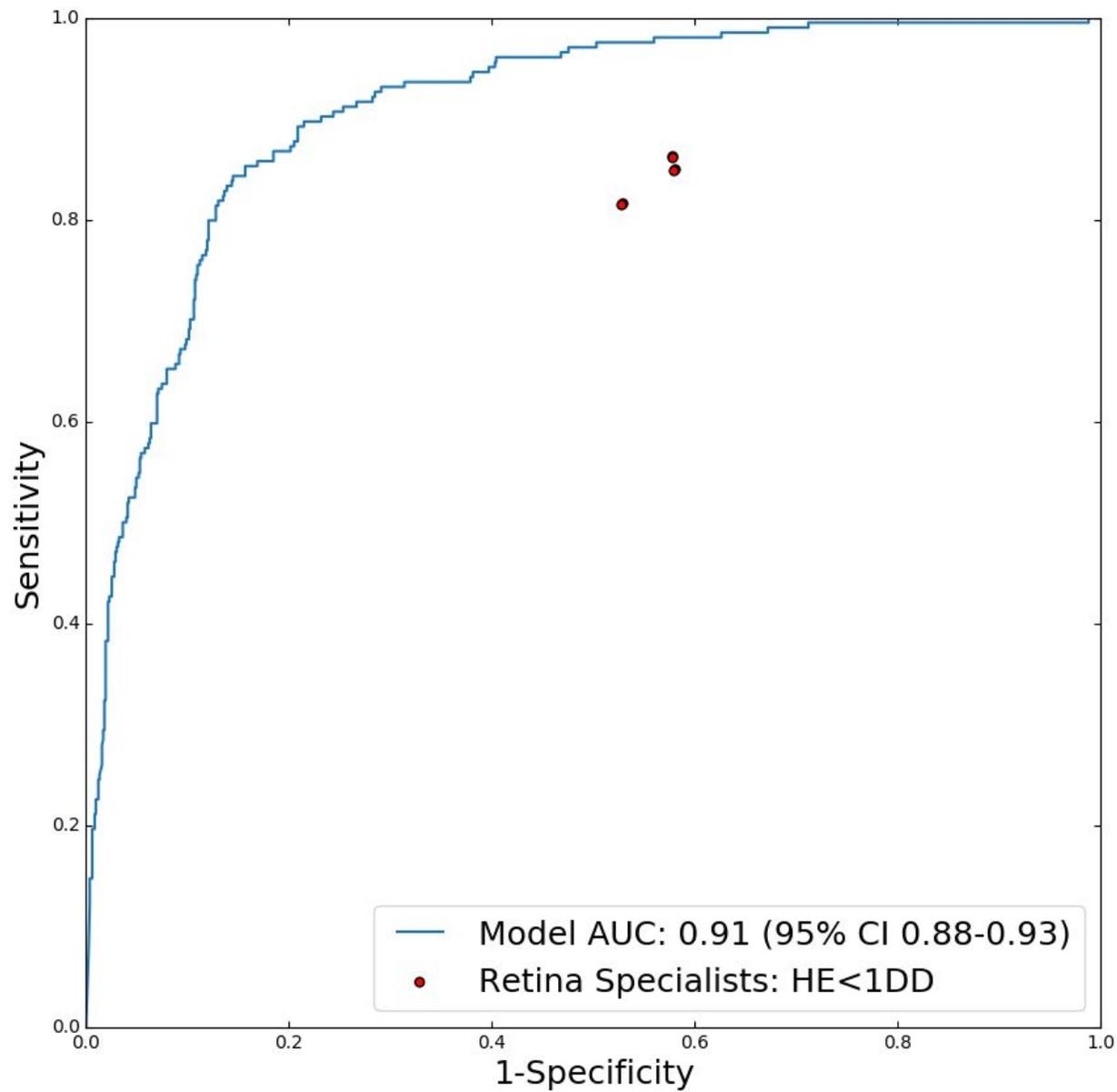

## C. CPT>=320 µm

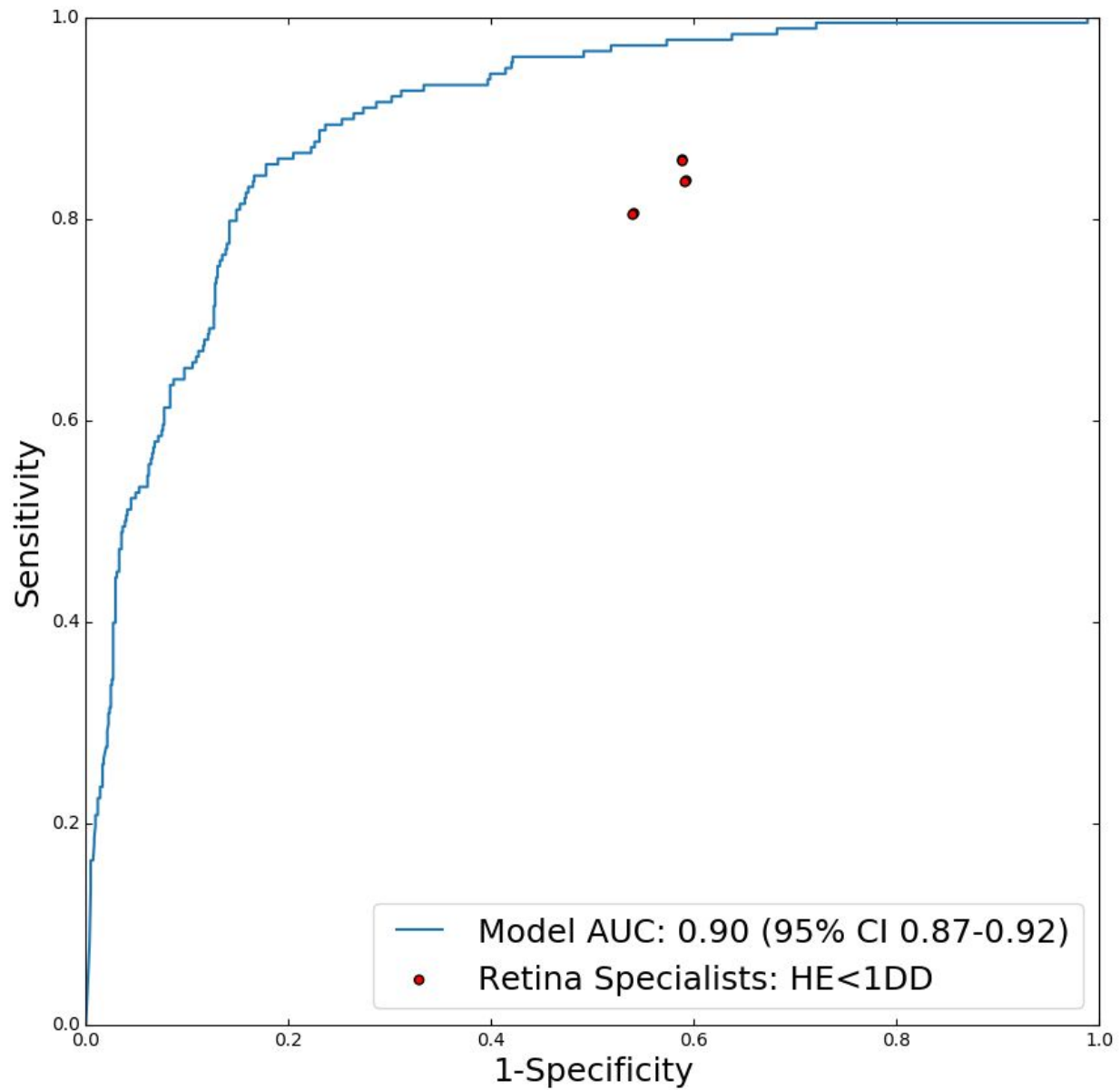

**Figure S2:** Receiver operating characteristic curve of the model with retinal specialists' grades shown as red dots for predicting ci-DME on the primary clinical validation set. All methods (i.e. the model and retinal specialists) rendered their grades using monoscopic fundus images only. The ground truth was derived using OCT at different center point thickness cut-offs for the definition of ci-DME. (A) 280 µm, (B) 300 µm, and (C) 320 µm.

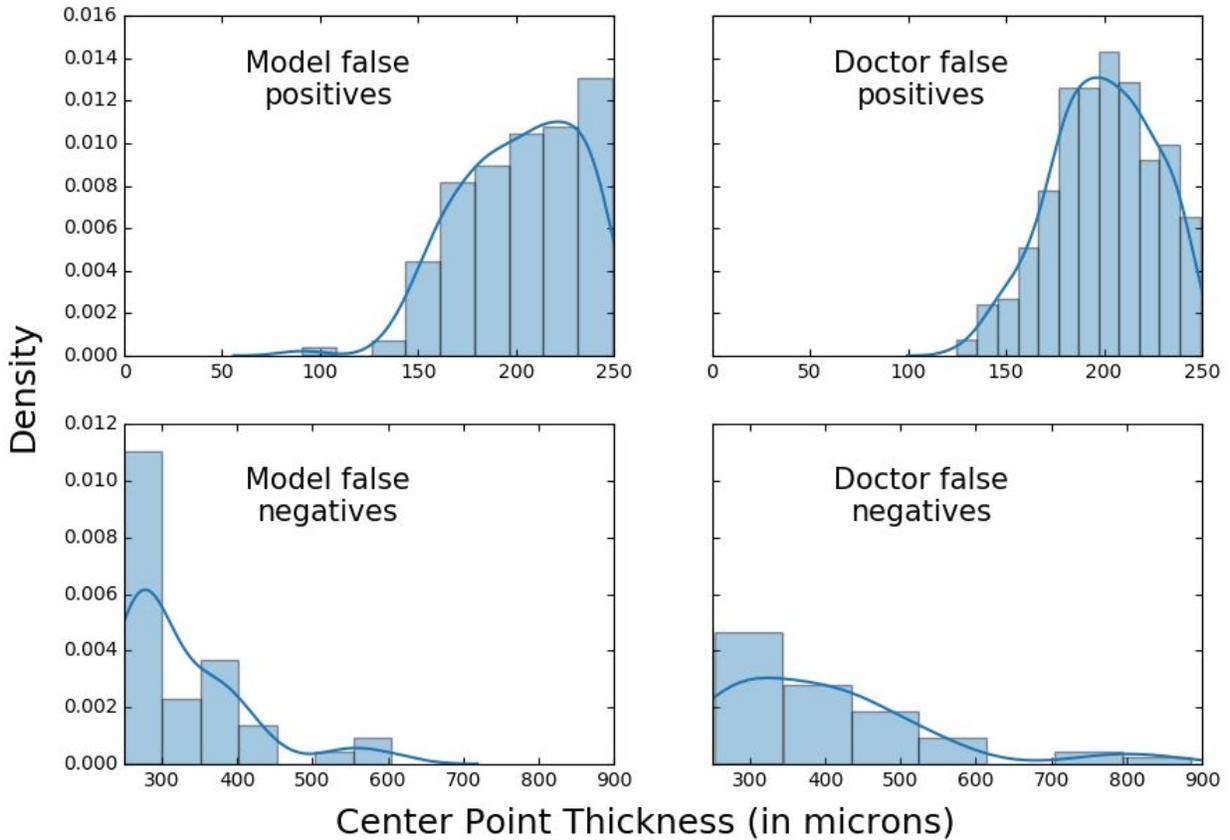

**Figure S3:** Center point thickness distribution of false positive (top) and false negative (bottom) instances for the model (left) and the retina specialists (right). To combine the grades from the three retina specialists we considered the case to be positive for ci-DME only if a majority of retina specialists agreed it was referable for DME. For the model, we chose an operating point that matched the sensitivity of the retina specialists.

- Inception-v3 architecture [See `tf.contrib.slim.nets.inception_v3`]. Weights initialized from a checkpoint trained for the Imagenet classification task.
- Input image resolution: 587 × 587
- Learning rate: 0.001
- Batch size: 32
- Weight decay: $4 \cdot 10^{-5}$
- Dropout keep probability: 0.8
- An Adam optimizer with $\beta_1 = 0.9$, $\beta_2 = 0.999$, and $\varepsilon = 0.1$ [see `tf.train.AdamOptimizer`]
- Data augmentation (in order):
    1. Random vertical and horizontal reflections [see `tf.image.random_flip_left_right` and `tf.image.random_flip_up_down`]
    2. Random brightness changes (with a max delta of 0.114752799273) [see the TensorFlow function `tf.image.random_brightness`]
    3. Random saturation changes between 0.559727311134 and 1.27488446236 [see `tf.image.random_saturation`]
    4. Random hue changes between -0.0251487996429 and 0.0251487996429 [see `tf.image.random_hue`]
    5. Random contrast changes between 0.999680697918 and 1.77048242092 [see `tf.image.random_contrast`]
- The model was trained for 2 million steps
- Model evaluations performed using a running average of parameters, with a decay factor of 0.9999

**Figure S4:** Data augmentation and model hyperparameters used for training the model.

| Metric | Model | Specialist 1 | Specialist 2 | Specialist 3 |
|---|---|---|---|---|
| Positive Predictive Value (%), 95% CI | 61% [57%-66%] | 36% [33%-40%] | 37% [33%-41%] | 38% [34%-42%] |
| Negative Predictive Value (%), 95% CI | 93% [91%-95%] | 88% [85%-92%] | 89% [86%-92%] | 87% [84%-90%] |
| Sensitivity (%), 95% CI | 85% [81%-89%] | 85% [80%-89%] | 86% [82%-90%] | 81% [76%-86%] |
| Specificity (%), 95% CI | 80% [77%-83%] | 44% [40%-47%] | 44% [40%-48%] | 49% [45%-53%] |
| Accuracy (%), 95% CI | 81% [78%-84%] | 55% [52%-58%] | 56% [53%-59%] | 57% [55%-61%] |
| Cohen's Kappa, 95% CI | 0.58 [0.52-0.63] | 0.20 [0.16-0.24] | 0.21 [0.17-0.25] | 0.22 [0.18-0.27] |

**Table S1:** Performance metrics (PPV, NPV, Sensitivity, Specificity, Accuracy and Cohen's Kappa) of the model for predicting ci-DME compared with the 3 retinal specialists, calculated only on the images that all 3 retinal specialists deemed gradable (n=948). For the model, we chose an operating point that matched the sensitivity of the retinal specialists to calculate the metrics. Brackets denote 95% confidence intervals.

| | **Thailand dataset** | **EyePACS-DME dataset** |
|---|---|---|
| Patient Population | Patients in Thailand presenting to a retina clinic of a tertiary hospital | Patients in a DR screening program determined based on CFP to have Moderate+ DR |
| OCT Device | Heidelberg Spectralis | Optovue iVue |
| ci-DME | Manual measurement of center point thickness >= 250um | Automated measurement of central subfield thickness from Optovue's software >= 300um |
| Cases with Epimacular Membrane | Excluded | Not excluded |
| Cases with macular edema from other causes | Excluded | Not excluded |
| Cases with proliferative DR with neovascular membrane affecting the macula | Excluded | Not excluded |
| Cases with previous laser treatment | Excluded | Not excluded |

**Table S2:** Comparison of the Thailand (training & primary validation) dataset and EyePACS-DME (secondary validation) dataset.